\documentclass[conference]{IEEEtran}
\IEEEoverridecommandlockouts
\usepackage{cite}
\usepackage{amsmath,amssymb,amsfonts}
\usepackage{algorithmic}
\usepackage{graphicx}
\usepackage{textcomp}
\usepackage{xcolor}
\usepackage{pifont}
\def\BibTeX{{\rm B\kern-.05em{\sc i\kern-.025em b}\kern-.08em
    T\kern-.1667em\lower.7ex\hbox{E}\kern-.125emX}}
\begin{document}

\title{Divide and Conquer in Video Anomaly Detection: A Comprehensive Review and New Approach\\

% \thanks{This work is supported by the National Natural Science Foundation of China (41971343).}
}

\author{\IEEEauthorblockN{Jian Xiao}
% \IEEEauthorblockA{\textit{School of Computer and Electronic Information} \\
\IEEEauthorblockA{\textit{School of Computer and Electronic} \\
\textit{Information /Artificial Intelligence} \\
\textit{Nanjing Normal University}\\
Nanjing, China \\
212202033@njnu.edu.cn}
\and
\IEEEauthorblockN{Tianyuan Liu}
\IEEEauthorblockA{\textit{Department of Industrial} \\
\textit{and Systems Engineering} \\
\textit{The Hong Kong Polytechnic University}\\
Hong Kong, China \\
tianyuan.liu@polyu.edu.hk}
\and
\IEEEauthorblockN{Genlin Ji*}
% \IEEEauthorblockA{\textit{School of Computer and Electronic Information} \\
\IEEEauthorblockA{\textit{School of Computer and Electronic} \\
\textit{Information /Artificial Intelligence} \\
\textit{Nanjing Normal University}\\
Nanjing, China \\
glji@njnu.edu.cn}
}
% \IEEEauthorblockA{\textit{School of Computer and Electronic Information} \\
\maketitle

\begin{abstract}
Video anomaly detection is a complex task, and the principle of "divide and conquer" is often regarded as an effective approach to tackling intricate issues. It's noteworthy that recent methods in video anomaly detection have revealed the application of the divide and conquer philosophy (albeit with distinct perspectives from traditional usage), yielding impressive outcomes. This paper systematically reviews these literatures from six dimensions, aiming to enhance the use of the divide and conquer strategy in video anomaly detection. Furthermore, based on the insights gained from this review, a novel approach is presented, which integrates human skeletal frameworks with video data analysis techniques. This method achieves state-of-the-art performance on the ShanghaiTech dataset, surpassing all existing advanced methods.
\end{abstract}


\section{Introduction}
Video anomaly detection (VAD) is an important problem in the field of computer vision. Its goal is to automatically identify anomalous events or behaviors within videos, such as sudden incidents, unusual objects, or atypical activities. Due to the rarity and diverse types of anomaly events compared to normal events, it is challenging for researchers to collect a large number of anomaly samples for traditional supervised training. Instead, they are constrained to employ easily obtainable normal data for model training.

This paper aims to investigate the application of the Divide and Conquer (DAC) approach in the domain of VAD. In recent years, the DAC approach as an effective method for solving complex problems has been found in a growing number of literature on VAD. The fundamental concept of the Divide and Conquer approach involves breaking down a problem into multiple independent sub-problems, solving each of these sub-problems separately, and then merging the results to attain the ultimate solution. From this, we can provide the following definition:

\noindent
\textbf{Definition 1.} \textit{In video anomaly detection, the Divide and Conquer approach refers to decomposing the problem along specific dimensions into multiple independent sub-problems. Each sub-problem independently provides anomaly scores/results based on its respective dimension, and these scores/results are then merged to obtain the overall anomaly score/result for the video.}

However, it is important to clarify that the Divide and Conquer approach in VAD, as studied in this paper, differs from traditional approaches, particularly in its perspective of partitioning subproblems. In traditional Divide and Conquer methods, the focus is often on the scale of the problem, addressing basic tasks such as search and sorting. However, VAD is a complex task that can be divided into subproblems from multiple perspectives, such as temporal and spatial dimensions, as well as different modalities. The former refers to the separation of the problem into temporal anomaly detection and spatial anomaly detection, with each sub-problem independently processing distinct information. The latter refers to the extraction of various modalities from the raw video data, such as RGB, optical flow, and human skeleton, treating the processing of each modality as a separate sub-problem. This situation corresponds to the later-stage fusion in multi-modal fusion\cite{baltruvsaitis2018multimodal}.

\begin{table*}[h!]
\caption{
Summary of Methods Employing Divide and Conquer Strategy}
\begin{center}
\scalebox{1.3}
{
\begin{tabular}{c c c c c c c c}
\hline
% \textbf{Table}&\multicolumn{3}{|c|}{\textbf{Table Column Head}} \\
% \cline{2-4} 
\textbf{Venue} & \textbf{InMod}& \textbf{TrainFocus}& \textbf{ModelProc}& \textbf{ModelBranch}& \textbf{OutMod}& \textbf{TestFocus} \\
\hline
CVPR 2019\cite{skeletonlearning} & 0 & 0 & 0 & 2 & 0 & 0 \\
\hline
ACM MM 2020\cite{cloze} & 0 & 0 & 0 & 2 & 1 & 0 \\
\hline
CVPR 2021\cite{multitask} & 0 & 0 & 1 & 4 & 0 & 0 \\
\hline
AAAI 2022\cite{bi-directional} & 0 & 0 & 0 & 2 & 1 & 0 \\
\hline
ACM MM 2022\cite{hierarchical} & 0 & 1 & 1 & 2 & 0 & 1 \\
\hline
arXiv 2022\cite{knowledge} & 0 & 0 & 1 & 2 & 0 & 0 \\
\hline
PR 2022\cite{implicit} & 0 & 0 & 1 & 3 & 0 & 0 \\
\hline
arXiv 2022\cite{attribute} & 1 & 0 & 1 & 3 & 0 & 0 \\
\hline
ECCV 2022\cite{jigsaw} & 0 & 0 & 0 & 2 & 0 & 0 \\
\hline
CVIU 2023\cite{ssmtl++} & 0 & 0 & 1 & 9 & 1 & 0 \\
\hline
\end{tabular}
}
\label{tab1}
\end{center}
\end{table*}

Observing the research in VAD in recent years, we have noticed an increasing number of papers achieving outstanding performance that demonstrate the utilization of the divide and conquer approach in their method design. Therefore, we argue that a reasonable application of the Divide and Conquer approach can enhance anomaly detection performance (This viewpoint is corroborated by the "ablation experiments" in the various references in Table \ref{tab1}). To explore more effective approaches for the application of the divide and conquer methodology, we conducted a review of literature that employs the Divide and Conquer approach, focusing on how it is utilized from different perspectives to improve anomaly detection performance. Recognizing the intricate nature of employing the divide and conquer approach in existing literature, we introduced a six-dimensional classification method that spans input data modality, training focal point, modeling process, modeling branches, output data modality, and testing focal point. This allowed us to effectively categorize these studies. Furthermore, based on the findings of our review, we posit that when partitioning subproblems from a modality perspective, the significant variations in data characteristics among different modalities should lead to distinct modeling techniques for each modality. By fully harnessing the inherent features of each modality's data and subsequently amalgamating their outcomes, the potential for achieving superior results could be enhanced. To validate this viewpoint, we introduced a novel method that integrates two representative existing approaches: one based on human skeleton information called STG-NF\cite{STG-NF}, and another based on RGB data known as Jigsaw\cite{jigsaw}. We conducted experiments on four datasets to comprehensively assess both the strengths and limitations of our proposed method.

% \begin{table*}[h!]
% \caption{Table Type Styles}
% \begin{center}
% \begin{tabular}{c c c c c c c c c}
% \hline
% % \textbf{Table}&\multicolumn{3}{|c|}{\textbf{Table Column Head}} \\
% % \cline{2-4} 
% \textbf{Venue} & \textbf{External pretrained models} & \textbf{InMod}& \textbf{InFocus}& \textbf{ModelProc}& \textbf{ModelBranch}& \textbf{OutMod}& \textbf{OutFocus} \\
% \hline
% CVPR 2019\cite{skeletonlearning} & AlphaPose$^{\mathrm{a}}$ & 0 & 0 & 0 & 2 & 0 & 0 \\
% \hline
% ACM MM 2020\cite{cloze} & Cascade R-CNN,FlowNet & 0 & 0 & 0 & 2 & 1 & 0 \\
% \hline
% CVPR 2021\cite{multitask} & YOLOv3,ResNet-50 & 0 & 0 & 1 & 4 & 0 & 0 \\
% \hline
% AAAI 2022\cite{bi-directional} & Cascade R-CNN,FlowNet2 & 0 & 0 & 0 & 2 & 1 & 0 \\
% \hline
% ACM MM 2022\cite{hierarchical} & Cascade R-CNN & 0 & 1 & 1 & 2 & 0 & 1 \\
% \hline
% arXiv 2022\cite{knowledge} & I3D & 0 & 0 & 1 & 2 & 0 & 0 \\
% \hline
% PR 2022\cite{implicit} & - & 0 & 0 & 1 & 3 & 0 & 0 \\
% \hline
% arXiv 2022\cite{attribute} & Mask-RCNN,FlowNet2,CLIP & 1 & 0 & 1 & 3 & 0 & 0 \\
% \hline
% ECCV 2022\cite{jigsaw} & YOLOv3 & 0 & 0 & 0 & 2 & 0 & 0 \\
% \hline
% CVIU 2023\cite{ssmtl++} & YOLOv5,ResNet-50,SelFlow,Mask-RCNN,UniPose & 0 & 0 & 1 & 9 & 1 & 0 \\
% \hline
% \multicolumn{4}{l}{$^{\mathrm{a}}$Sample of a Table footnote.}
% \end{tabular}
% \label{tab1}
% \end{center}
% \end{table*}

This paper's contributions are summarized as follows:
\begin{itemize}
\item Establishing the definition of the divide and conquer approach within the realm of VAD, and conducting a comprehensive review of VAD literature that demonstrates the application of this approach from six different dimensions.
\item Drawing upon the results of the review, proposing the perspective of "potentially achieving superior outcomes by aggregating results from different modalities while fully harnessing their individual data" from the standpoint of modality-based division.
\item To validate the aforementioned perspective, devising a novel method that integrates techniques rooted in human body skeletal information and RGB data. This method attains an AUC score of 87.72 on the ShanghaiTech\cite{shanghaitech} dataset, surpassing all current state-of-the-art methods.
\end{itemize}

\section{Taxonomy}
Given the complexity of the video anomaly detection task, different methods showcase distinct strategies when employing the divide and conquer approach. In order to systematically categorize the divide and conquer strategies within existing methods, we introduce a six-dimensional classification scheme.
\begin{itemize}
\item Input Data Modality: Whether the input data contains multiple modalities.
\item Training Focus: Whether the training data consists of different entities.
\item Modeling Process: Whether significantly different modeling techniques are employed for modeling the input data.
\item Modeling Branches: The number of branches in the modeling process.
\item Output Data Modality: Whether the output data contains multiple modalities.
\item Testing Focus: Whether the testing data involves different entities.
\end{itemize}
The summary results are shown in Table \ref{tab1}, where apart from the Modeling Branches dimension which takes values from the set of natural numbers, the other dimensions have values of either 0 or 1. In this context, 0 signifies negation, akin to "No," while 1 represents affirmation, akin to "Yes."

From Table \ref{tab1}, it is evident that there is currently a scarcity of literature on dividing problems into sub-problems based on data modalities or according to training-test focal points, with only references\cite{attribute} and\cite{hierarchical} available, respectively. Moreover, we note that the method\cite{attribute}, which employs a divide-and-conquer strategy based on data modality, achieves the highest or near-highest performance on three commonly used datasets, as shown in Table \ref{tab3}. Upon analyzing\cite{attribute}, it becomes apparent that it doesn't utilize overly complex modeling techniques, relying solely on simple methods such as Gaussian Mixture Models (GMM) and K-Nearest Neighbors (KNN) for modeling. As a result, we attribute its performance improvement to the use of the modality divide-and-conquer method. Different data modalities exhibit varying degrees of sensitivity to anomalies (levels of difficulty in capturing anomalies). As long as anomalies that are easy to detect within each respective modality are identified, fusing the anomaly detection outcomes from different modalities can lead to a more comprehensive anomaly detection outcome. Based on this, we believe that by thoroughly exploring the single-modal data using meticulously designed models, there exists the potential to further enhance the anomaly detection performance of the model. To validate this hypothesis, we adopt a modality-based divide-and-conquer approach, proposing a novel method that combines the human skeletal-based approach with the RGB data-based approach.

\section{Approach}

\begin{figure*}[htbp]
\includegraphics[width=\linewidth]{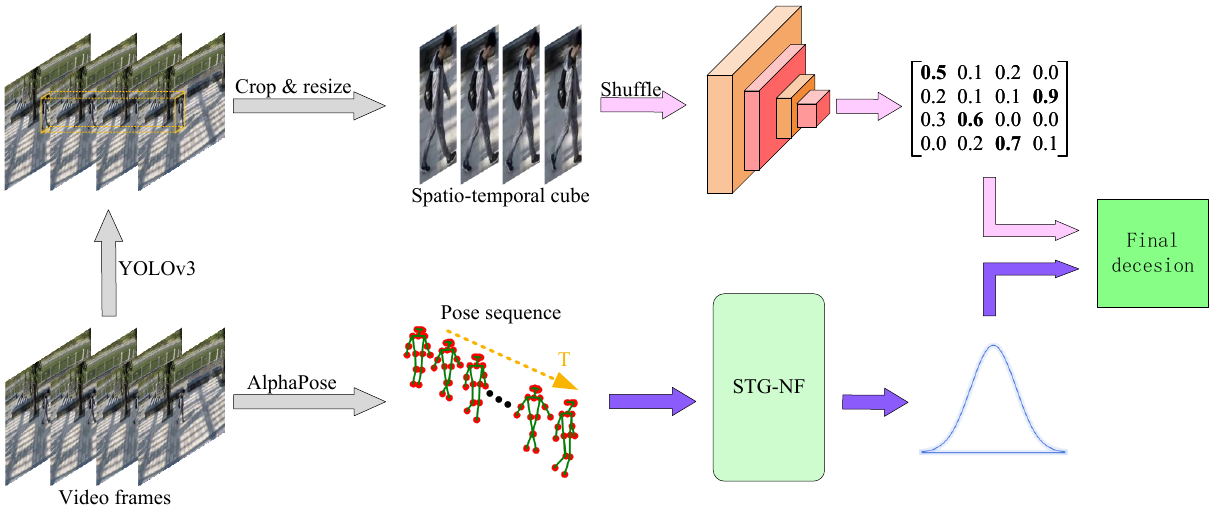}
\caption{Method overview: The approach consists of two branches. One branch employs video data processed through a target detection mechanism for anomaly identification, while the other branch leverages skeletal data modeling for anomaly detection. Ultimately, the results from both branches are ensemble-aggregated to derive the conclusive anomaly detection outcomes.}
\label{framework}
\end{figure*}

\subsection{Overview}
To validate the viewpoints presented in this article, we divide the problem from a modal perspective, categorizing it into two sub-problems: anomaly detection in videos based on human skeletal structure and anomaly detection in videos based on RGB data. In order to fully exploit the potential of single-modal data, we have chosen two methods that have demonstrated excellent performance in recent years: Jigsaw\cite{jigsaw}, which relies on RGB data, and STG-NF\cite{STG-NF}, which operates on skeletal data. Due to space constraints, we provide a general overview of these two methods. The overall framework is illustrated in Figure\ref{framework}.

\subsection{RGB-based Approach}
This method employs a normalizing flow to estimate the density of human skeletal data, yielding log-likelihood scores used for anomaly detection. The underlying assumption is that normal instances reside within high-density regions, while anomalies are found in low-density areas. The method involves three steps: 1) Skeleton Sequence Extraction: Given a video, start by extracting skeletal data using AlphaPose\cite{li2019alphapose}. Then, apply OSNet\cite{osnet} tracking to obtain a sequence of human skeletal data. Finally, divide this sequence into fixed-time segments. 2) Modeling Skeleton Sequences: Feed these skeletal sequence segments into Spatio-Temporal Graph Normalizing Flows for direct modeling and outputting corresponding log-likelihood scores. 3) Obtaining Anomaly Scores: Choose the minimum log-likelihood score from each skeletal sequence segment within every frame as the anomaly score for that frame. For further details, please refer to the reference\cite{STG-NF}.

\subsection{Skeleton-based Approach}
This approach employs a self-supervised learning framework for anomaly detection by solving a spatio-temporal jigsaw puzzle. The method involves three steps: 1) Constructing Spatio-Temporal Cubes: Given a video, first utilize YOLOv3\cite{yolov3} for foreground object detection in each frame, generating a series of bounding boxes. Then, for each detected object within a frame and its neighboring frames, extract a sequence of image patches based on the object's bounding box. These image patches are resized to a fixed size and stacked along the temporal dimension, forming spatio-temporal cubes centered around the object. 2) Solving the Puzzle: Shuffle the temporal or spatial dimensions of each spatio-temporal cube and feed them into a fully convolutional network to reconstruct the correct sequence of the cubes. 3) Obtaining Anomaly Scores: Choose the minimum anomaly score from each spatio-temporal cube within every frame as the anomaly score for that frame. For more details, please refer to the reference\cite{jigsaw}.

\subsection{Anomaly Scores}
Due to skeleton-based methods only detecting human-related anomalies and RGB-based methods detecting both human-related and unrelated anomalies, we designed two fusion approaches (Depending on the varying focus of the tests):
\begin{itemize}
\item STG-NF detects human-related anomalies, while Jigsaw detects non-human-related anomalies.
\item STG-NF detects human-related anomalies, and Jigsaw detects both human-related and non-human-related anomalies.
\end{itemize}

Specifically, it means that when Jigsaw detects only non-human-related anomalies, the bounding boxes of humans are removed from the test dataset. After obtaining anomaly scores from the two branches, their values are summed to yield the final anomaly score.

\begin{figure}[htbp]
\includegraphics[width=\linewidth]{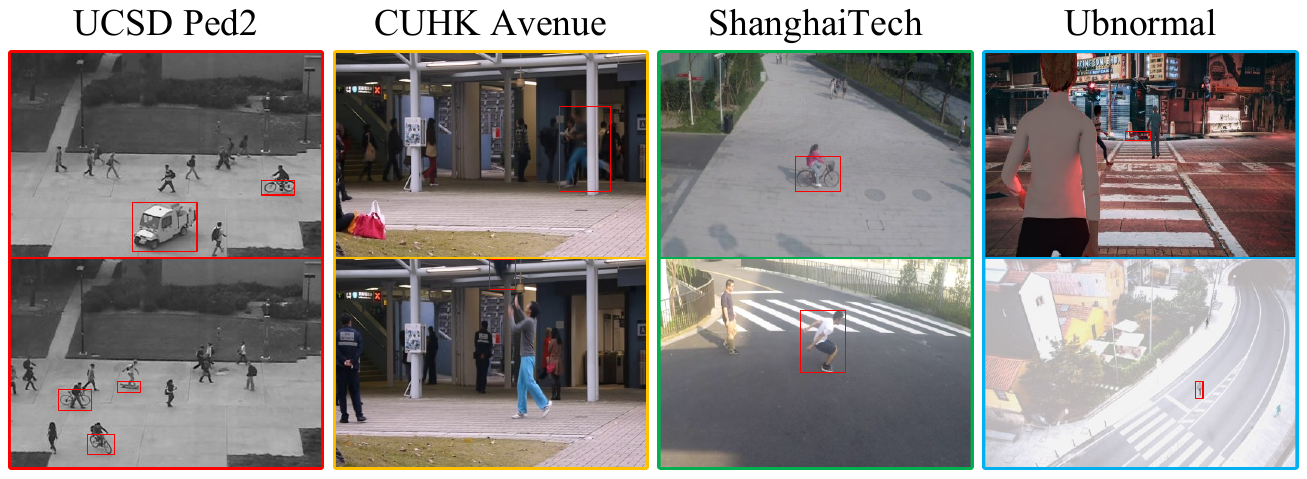}
\caption{Displayed are samples of abnormal frames from the UCSD Ped2, CUHK Avenue, ShanghaiTech, and UBnormal datasets, with anomalies highlighted by red boxes.}
\label{dataset}
\end{figure}

\section{Experiments}

\subsection{Experimental Setting}
\paragraph{Datasets}
To comprehensively test the performance of our method, we selected four datasets with distinct features for experimentation. Some samples are shown in Figure\ref{dataset}.

\textbf{UCSD Ped2}\cite{ped2}: This single-scene pedestrian dataset captures distant camera shots in dense crowds, featuring anomalies like skateboarding, cycling, and large trucks. The unique characteristics include small-sized targets and a crowded environment.

\textbf{CUHK Avenue}\cite{avenue}: Set in a campus environment, this single-scene dataset captures anomalies such as sudden running, bag throwing, and playful behavior. The dataset's notable aspects are frontal camera angles, relatively dense crowds, and significant occlusions.

\textbf{ShanghaiTech}\cite{shanghaitech}: A multi-scene campus dataset encompassing 13 different scenarios, it includes both human-related anomalies like jumping and cycling, as well as non-human-related anomalies like large vehicle presence. Noteworthy features include diverse scenes, angles, lighting conditions, and fewer instances of crowd occlusion.

\textbf{UBnormal}\cite{ubnormal}: This synthetically generated multi-scene dataset comprises 29 scenes and encompasses 22 anomaly types with 660 events. The dataset poses highly challenging anomalies like traffic rule violations, sleeping, and theft. Its distinctiveness lies in the diversity of scenes (including foggy and smoky conditions) and a wide range of anomaly categories.

\paragraph{Evaluation Metrics}
To assess the model's performance, this study employs frame-level Micro-AUC\cite{georgescu2021background} (Area Under the ROC Curve) as the primary measure. Specifically, this involves sequentially concatenating frames from all videos in the test set and calculating the AUC. A higher value indicates superior model performance.

\begin{table}[h]
\caption{Utilization of Datasets for Jigsaw and STG-NF in the Original Papers}
\begin{center}
\scalebox{0.87}
{
\begin{tabular}{c c c c c}
\hline
\quad & \textbf{UCSD Ped2} & \textbf{CUHK Avenue}& \textbf{ShanghaiTech}& \textbf{UBnormal} \\
\hline
Jigsaw\cite{jigsaw} & \ding{52} & \ding{52} & \ding{52} & \quad\\
STG-NF\cite{STG-NF} & \quad & \quad & \ding{52} & \ding{52} \\
\hline
\end{tabular}
}
\label{tab2}
\end{center}
\end{table}

\begin{figure*}[htbp]
\includegraphics[width=\linewidth]{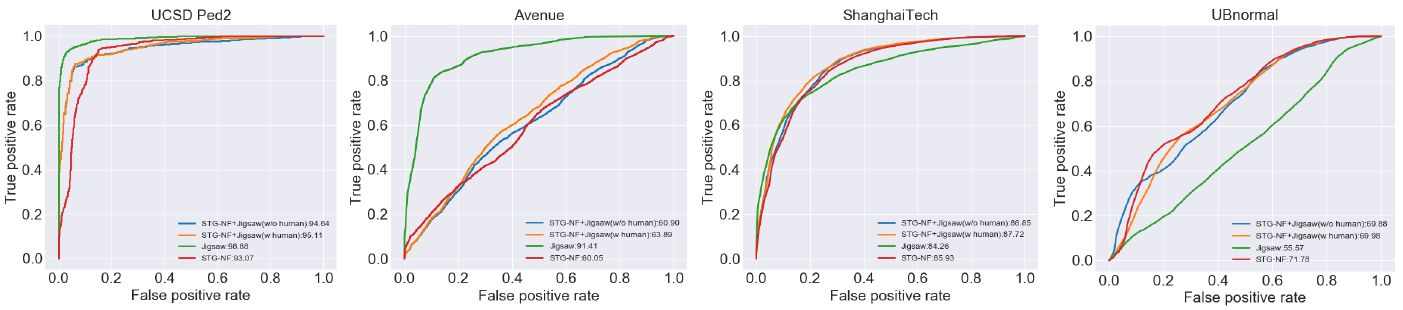}
\caption{Comparison of frame-level ROC curves on different VAD benchmark datasets.}
\label{roc}
\end{figure*}

\begin{table*}[h]
\caption{Comparison with State-of-the-Art Methods on Different VAD benchmark datasets}
\begin{center}
\scalebox{1.1}
{
\begin{tabular}{c c c c c c}
\hline
% \textbf{Table}&\multicolumn{3}{|c|}{\textbf{Table Column Head}} \\
% \cline{2-4} 
\textbf{Year} & \textbf{Method} & \textbf{UCSD Ped2} & \textbf{CUHK Avenue}& \textbf{ShanghaiTech}& \textbf{UBnormal} \\
\hline
2020 & Cloze Test\cite{cloze} & 97.3 & 89.6 & 74.8 & - \\
2020 & Background-Agnostic Framework\cite{georgescu2021background} & 98.7 & 92.3 & 82.7 & - \\
2021 & HF$^{2}$-VAD\cite{liu2021hybrid} & 99.3 & 91.1 & 76.2 & - \\
2021 & SSMTL\cite{multitask} & - & 91.5 & 82.4 & - \\
2022 & HSNBM\cite{hierarchical} & 95.2 & 91.6 & 76.5 & - \\
2022 & ITAE + NFs\cite{implicit} & 99.2 & 88.0 & 76.3 & - \\
2022 & HF$^{2}$-VAD + SSPCAB\cite{2022SSPCAB} & - & 90.9 & 75.5 & - \\
2022 & Bi-Directional VAD Framework\cite{bi-directional} & 98.3 & 90.3 & 78.1 & - \\
2022 & SSMTL++v1\cite{ssmtl++} & - & \textbf{93.7} & 82.9 & 62.1 \\
2022 & Two-Stream\cite{knowledge} & 97.1 & 90.8 & 83.7 & - \\
2022 & Background-Agnostic Framework + SSPCAB\cite{2022SSPCAB} & - & 92.9 & 83.6 & - \\
2022 & Jigsaw\textbf{*}\cite{jigsaw} & 98.88 & 91.41 & 84.26 & 55.57 \\
2022 & STG-NF\textbf{*}\cite{STG-NF} & 93.07 & 60.05 & 85.93 & \textbf{71.78} \\
2023 & AMSRC\cite{huang2023amsrc} & \textbf{99.5} & \textbf{93.8} & 76.6 & - \\
2023 & EVAL\cite{singh2023eval} & - & 86.02 & 76.63 & - \\
2023 & DMAD\cite{liu2023diversity} & \textbf{99.7} & 92.8 & 78.8 & - \\
2023 & SSMTL++v2\cite{ssmtl++} & - & 91.6 & 83.8 & 56.0 \\
2023 & AI-VAD\cite{attribute} & 99.1 & 93.6 & 85.94 & - \\
\hline
2023 & STG-NF + Jigsaw(w/o human) & 94.64 & 60.90 & \textbf{86.85} & 69.88 \\
2023 & STG-NF + Jigsaw(w human) & 95.11 & 63.89 & \textbf{87.72} & \textbf{69.98} \\
\hline
\multicolumn{4}{l}{$^{*}$denotes that the results in the table are from our implementation.}
\end{tabular}
}
\label{tab3}
\end{center}
\end{table*}

\paragraph{Implementation Details}
As shown in Table \ref{tab2}, both Jigsaw and STG-NF papers do not provide complete results across all four datasets. Therefore, for the missing results, we conducted supplementary experiments using their open-source code, and the outcomes are presented in Table \ref{tab3}. For the experiments on the UBnormal dataset, Jigsaw was tested with a filter setting of 0.8 and a sample length of 9. STG-NF conducted experiments on the Ped2 and Avenue datasets, with segmentation lengths set at 24 for both. Other parameters remained consistent with the original papers, as referenced in\cite{jigsaw} and\cite{STG-NF}. In cases where humans were absent, the anomaly scores were set to 0, such as in the last two test videos of Ped2.

\subsection{Experimental Results and Analysis}
We conducted a comprehensive comparison between the method proposed in this paper and eighteen other state-of-the-art approaches, regardless of their adoption of the divide-and-conquer strategy. The comparative results are presented in Table \ref{tab3}, with the top two performances being highlighted in bold. Additionally, we have depicted frame-level ROC curves in Figure \ref{roc}. Notably, the method introduced in this study exhibits a significant performance enhancement on the ShanghaiTech dataset in comparison to all other advanced techniques, achieving an impressive AUC of 87.73\%. To the best of our knowledge, this represents the most remarkable performance ever achieved on the ShanghaiTech dataset. Nevertheless, it is imperative to acknowledge that, when contrasted with the pre-fused Jigsaw and STG-NF methods, the fused approach demonstrates performance superiority exclusively on the ShanghaiTech dataset. On the remaining three datasets (Ped2, Avenue, and UBnormal), the fused method falls short of surpassing the original performance. Digging deeper into the underlying reasons, we posit that this observed phenomenon precisely aligns with our conjecture. When distinct modeling approaches are applied to different modalities of data, the potential for achieving enhanced results through modal aggregation arises only when a thorough exploration of each modality's information is undertaken. Should at least one method exhibit subpar performance, the subsequent fusion of outcomes will likely be sub-optimal. To elaborate, the reason the merged results could outperform on the ShanghaiTech dataset is rooted in the commendable individual performances of the pre-merged methods—Jigsaw and STG-NF—registering 84.26\% and 85.93\% respectively. In contrast, on other datasets, instances of notably poor performance by either Jigsaw or STG-NF exist, such as STG-NF's 93.07\% on Ped2 and 60.05\% on Avenue, or Jigsaw's 55.57\% on UBnormal.

\begin{figure*}[htbp]
\begin{center}
\scalebox{0.9}{
\includegraphics[width=\linewidth]{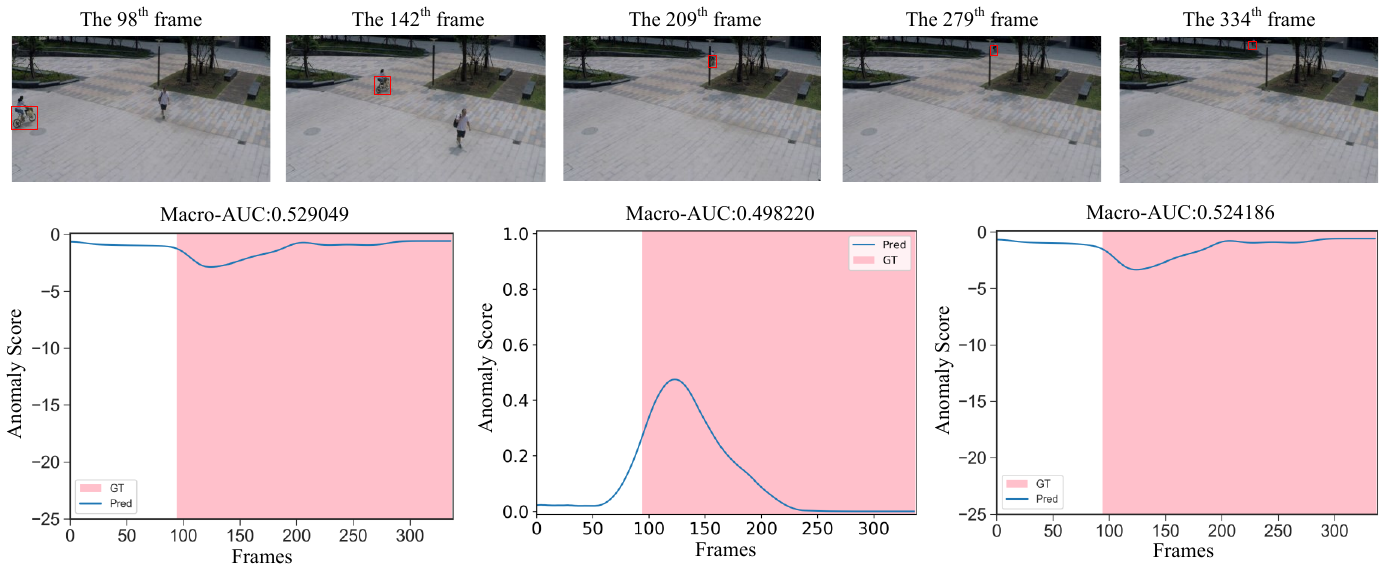}
}
\end{center}
\caption{Visualization of Predictive Results for STG-NF\cite{STG-NF}, Jigsaw\cite{jigsaw}, and Proposed Method on ShanghaiTech Dataset.}
\label{visualization}
\end{figure*}

The substantial performance disparities between the two methods are largely attributed to the characteristics of the datasets, as elucidated in the dataset section. In the Ped2 and Avenue datasets, the dense crowd scenarios and substantial occlusions hinder the successful detection and tracking of human skeletons. On the UBnormal dataset, the dynamic scenes and rich semantic information of anomaly types render the effective detection of anomalies challenging without relying on coarse-grained visual cues. Furthermore, we visualize the instances of poor performance for the three methods on the ShanghaiTech dataset, as depicted in Figure \ref{visualization}. The top row illustrates key frames from the respective videos, while the bottom row, from left to right, showcases the anomaly detection outcomes of STG-NF, Jigsaw, and our proposed method on the corresponding videos. From these illustrations, it becomes evident that once anomalies become subtle, all three methods falter in detection. This highlights that the divide-and-conquer strategy employed in our method is not optimal, necessitating exploration of superior modeling techniques and partitioning strategies.

Lastly, upon comparing the two fusion approaches devised in this study, it becomes evident that across all datasets, Jigsaw's performance in detecting all anomalies surpasses its performance in solely detecting non-human-related anomalies. This implies that for anomalies involving humans, the fusion of STG-NF and Jigsaw detection outcomes results in complementary information. However, it also signifies that STG-NF's exploration of skeletal data is not exhaustive.

\section{Conclusion}
In this paper, we comprehensively review recent studies in the field of video anomaly detection that have employed a divide-and-conquer approach from six perspectives. Based on this, we propose the viewpoint that "distinct modalities with significant feature differences should be modeled differently, and by fully exploiting individual modality information, aggregating the results of each modality should yield superior outcomes." To validate this, experiments are conducted on four datasets, and this viewpoint receives preliminary confirmation on the ShanghaiTech dataset, where the proposed method achieves state-of-the-art performance. Note that our main contribution does not come from proposing a completely new method, but rather from proposing a new perspective on VAD—taking a divide-and-conquer approach. We sincerely hope our work can benefit future research in this area.

\section*{Acknowledgment}
This work is supported by the National Natural Science Foundation of China under Grants No.41971343.
\bibliography{xiao}

\begin{thebibliography}{10}

\bibitem{baltruvsaitis2018multimodal}
T.~Baltru{\v{s}}aitis, C.~Ahuja, and L.-P. Morency, ``Multimodal machine
  learning: A survey and taxonomy,'' {\em IEEE transactions on pattern analysis
  and machine intelligence}, vol.~41, no.~2, pp.~423--443, 2018.

\bibitem{skeletonlearning}
R.~Morais, V.~Le, T.~Tran, B.~Saha, M.~Mansour, and S.~Venkatesh, ``Learning
  regularity in skeleton trajectories for anomaly detection in videos,'' in
  {\em Proceedings of the IEEE/CVF conference on computer vision and pattern
  recognition}, pp.~11996--12004, 2019.

\bibitem{cloze}
G.~Yu, S.~Wang, Z.~Cai, E.~Zhu, C.~Xu, J.~Yin, and M.~Kloft, ``Cloze test
  helps: Effective video anomaly detection via learning to complete video
  events,'' in {\em Proceedings of the 28th ACM International Conference on
  Multimedia}, pp.~583--591, 2020.

\bibitem{multitask}
M.-I. Georgescu, A.~Barbalau, R.~T. Ionescu, F.~S. Khan, M.~Popescu, and
  M.~Shah, ``Anomaly detection in video via self-supervised and multi-task
  learning,'' in {\em Proceedings of the IEEE/CVF conference on computer vision
  and pattern recognition}, pp.~12742--12752, 2021.

\bibitem{bi-directional}
C.~Chen, Y.~Xie, S.~Lin, A.~Yao, G.~Jiang, W.~Zhang, Y.~Qu, R.~Qiao, B.~Ren,
  and L.~Ma, ``Comprehensive regularization in a bi-directional predictive
  network for video anomaly detection,'' in {\em Proceedings of the AAAI
  Conference on Artificial Intelligence}, vol.~36, pp.~230--238, 2022.

\bibitem{hierarchical}
Q.~Bao, F.~Liu, Y.~Liu, L.~Jiao, X.~Liu, and L.~Li, ``Hierarchical scene
  normality-binding modeling for anomaly detection in surveillance videos,'' in
  {\em Proceedings of the 30th ACM International Conference on Multimedia},
  pp.~6103--6112, 2022.

\bibitem{knowledge}
C.~Cao, Y.~Lu, and Y.~Zhang, ``Context recovery and knowledge retrieval: A
  novel two-stream framework for video anomaly detection,'' {\em arXiv preprint
  arXiv:2209.02899}, 2022.

\bibitem{implicit}
M.~Cho, T.~Kim, W.~J. Kim, S.~Cho, and S.~Lee, ``Unsupervised video anomaly
  detection via normalizing flows with implicit latent features,'' {\em Pattern
  Recognition}, vol.~129, p.~108703, 2022.

\bibitem{attribute}
T.~Reiss and Y.~Hoshen, ``Attribute-based representations for accurate and
  interpretable video anomaly detection,'' {\em arXiv preprint
  arXiv:2212.00789}, 2022.

\bibitem{jigsaw}
G.~Wang, Y.~Wang, J.~Qin, D.~Zhang, X.~Bao, and D.~Huang, ``Video anomaly
  detection by solving decoupled spatio-temporal jigsaw puzzles,'' in {\em
  European Conference on Computer Vision}, pp.~494--511, Springer, 2022.

\bibitem{ssmtl++}
A.~Barbalau, R.~T. Ionescu, M.-I. Georgescu, J.~Dueholm, B.~Ramachandra,
  K.~Nasrollahi, F.~S. Khan, T.~B. Moeslund, and M.~Shah, ``Ssmtl++: Revisiting
  self-supervised multi-task learning for video anomaly detection,'' {\em
  Computer Vision and Image Understanding}, vol.~229, p.~103656, 2023.

\bibitem{STG-NF}
O.~Hirschorn and S.~Avidan, ``Normalizing flows for human pose anomaly
  detection,'' {\em arXiv preprint arXiv:2211.10946}, 2022.

\bibitem{shanghaitech}
W.~Luo, W.~Liu, and S.~Gao, ``A revisit of sparse coding based anomaly
  detection in stacked rnn framework,'' in {\em Proceedings of the IEEE
  international conference on computer vision}, pp.~341--349, 2017.

\bibitem{li2019alphapose}
J.~Li, C.~Wang, H.~Zhu, Y.~Mao, H.-S. Fang, and C.~Lu, ``Crowdpose: Efficient
  crowded scenes pose estimation and a new benchmark,'' in {\em Proceedings of
  the IEEE/CVF conference on computer vision and pattern recognition},
  pp.~10863--10872, 2019.

\bibitem{osnet}
K.~Zhou, Y.~Yang, A.~Cavallaro, and T.~Xiang, ``Learning generalisable
  omni-scale representations for person re-identification,'' {\em IEEE
  transactions on pattern analysis and machine intelligence}, vol.~44, no.~9,
  pp.~5056--5069, 2021.

\bibitem{yolov3}
J.~Redmon and A.~Farhadi, ``Yolov3: An incremental improvement,'' {\em arXiv
  preprint arXiv:1804.02767}, 2018.

\bibitem{ped2}
V.~Mahadevan, W.~Li, V.~Bhalodia, and N.~Vasconcelos, ``Anomaly detection in
  crowded scenes,'' in {\em The Twenty-Third {IEEE} Conference on Computer
  Vision and Pattern Recognition, {CVPR} 2010, San Francisco, CA, USA, 13-18
  June 2010}, pp.~1975--1981, {IEEE} Computer Society, 2010.

\bibitem{avenue}
C.~Lu, J.~Shi, and J.~Jia, ``Abnormal event detection at 150 fps in matlab,''
  in {\em Proceedings of the IEEE international conference on computer vision},
  pp.~2720--2727, 2013.

\bibitem{ubnormal}
A.~Acsintoae, A.~Florescu, M.-I. Georgescu, T.~Mare, P.~Sumedrea, R.~T.
  Ionescu, F.~S. Khan, and M.~Shah, ``Ubnormal: New benchmark for supervised
  open-set video anomaly detection,'' in {\em Proceedings of the IEEE/CVF
  Conference on Computer Vision and Pattern Recognition}, pp.~20143--20153,
  2022.

\bibitem{georgescu2021background}
M.~I. Georgescu, R.~T. Ionescu, F.~S. Khan, M.~Popescu, and M.~Shah, ``A
  background-agnostic framework with adversarial training for abnormal event
  detection in video,'' {\em IEEE transactions on pattern analysis and machine
  intelligence}, vol.~44, no.~9, pp.~4505--4523, 2021.

\bibitem{liu2021hybrid}
Z.~Liu, Y.~Nie, C.~Long, Q.~Zhang, and G.~Li, ``A hybrid video anomaly
  detection framework via memory-augmented flow reconstruction and flow-guided
  frame prediction,'' in {\em Proceedings of the IEEE/CVF international
  conference on computer vision}, pp.~13588--13597, 2021.

\bibitem{2022SSPCAB}
N.-C. Ristea, N.~Madan, R.~T. Ionescu, K.~Nasrollahi, F.~S. Khan, T.~B.
  Moeslund, and M.~Shah, ``Self-supervised predictive convolutional attentive
  block for anomaly detection,'' in {\em Proceedings of the IEEE/CVF conference
  on computer vision and pattern recognition}, pp.~13576--13586, 2022.

\bibitem{huang2023amsrc}
X.~Huang, C.~Zhao, and Z.~Wu, ``A video anomaly detection framework based on
  appearance-motion semantics representation consistency,'' in {\em ICASSP
  2023-2023 IEEE International Conference on Acoustics, Speech and Signal
  Processing (ICASSP)}, pp.~1--5, IEEE, 2023.

\bibitem{singh2023eval}
A.~Singh, M.~J. Jones, and E.~G. Learned-Miller, ``Eval: Explainable video
  anomaly localization,'' in {\em Proceedings of the IEEE/CVF Conference on
  Computer Vision and Pattern Recognition}, pp.~18717--18726, 2023.

\bibitem{liu2023diversity}
W.~Liu, H.~Chang, B.~Ma, S.~Shan, and X.~Chen, ``Diversity-measurable anomaly
  detection,'' in {\em Proceedings of the IEEE/CVF Conference on Computer
  Vision and Pattern Recognition}, pp.~12147--12156, 2023.

\end{thebibliography}
\end{document}